\documentclass[runningheads]{llncs}
\usepackage{graphicx}
\usepackage{algorithm}
\usepackage{algorithmicx}
\newcommand{\specialcell}[2][c]{%
\begin{tabular}[#1]{@{}c@{}}#2\end{tabular}}

\usepackage{lipsum}
\usepackage{mwe}
\usepackage{amsmath}
\usepackage{booktabs}

\begin{document}
\title{Domain Adaptive Transfer Learning on Visual
Attention Aware Data Augmentation for Fine-grained
Visual Categorization}
\titlerunning{DATL on Visual Attention Aware Data Augmentation for FGVC}

\author{Ashiq Imran, \and
Vassilis Athitsos}

\authorrunning{Ashiq Imran, and Vassilis Athitsos}

\institute{Department of Computer Science and Engineering \\ University of Texas at Arlington, Arlington, TX, USA \\
\email{ashiq.imran@mavs.uta.edu}\\
\email{athitsos@uta.edu}}

\maketitle             

\begin{abstract}
Fine-Grained Visual Categorization (FGVC) is a challenging topic in computer vision. It is a problem characterized by large intra-class differences and subtle inter-class differences. In this paper, we tackle this problem in a weakly supervised manner, where neural network models are
getting fed with additional data using a data augmentation technique through a visual attention mechanism. We perform domain adaptive knowledge transfer via fine-tuning on our base network model. We perform our experiment on six challenging and commonly used FGVC datasets, and we show competitive improvement on accuracies by using attention-aware data augmentation techniques with features derived from deep learning model InceptionV3, pre-trained on large scale datasets. Our method outperforms competitor methods on multiple FGVC datasets and showed competitive results on other datasets. Experimental studies show that transfer learning from large scale datasets can be utilized effectively with visual attention based data augmentation, which can obtain state-of-the-art results on several FGVC datasets. We present a comprehensive analysis of our experiments. Our method achieves state-of-the-art results in multiple fine-grained classification datasets including challenging CUB200-2011 bird, Flowers-102, and FGVC-Aircrafts datasets.

\end{abstract}
\section{Introduction}
Deep neural networks have provided state-of-the-art results in many domains in computer vision. However, having a big training set is very important for the performance of deep neural networks \cite{deng2009imagenet,krizhevsky2012imagenet}. Data augmentation techniques have been gaining popularity in deep learning and are extensively used to address the scarcity of training data. Data augmentation has led to promising results in various computer vision tasks \cite{krizhevsky2012imagenet}. There are different data augmentation methods for deep models, like image flipping, cropping, scaling, rotation, translation, color distortion, adding Gaussian noise, and many more.

Previous methods mostly choose random images from the dataset and apply the above operations to enlarge the amount of training data. However, applying random cropping to generate new training examples can have undesirable consequences. For example, if the size of the cropped region is not large enough, it may consist entirely of background, and not contain any part of the labeled object. Moreover, this generated data might reduce accuracy and negatively affect the quality of the extracted features. Consequently, the disadvantages of random cropping might cancel out its advantages. More specific features need to be provided to the model to make data augmentation more productive. 

In Fine-Grained Visual Categorization (FGVC), same-class items may have variation in the pose, scale, or rotation. FGVC contains subtle differences among classes in a sub-category of an object, which includes the model of the cars, type of the foods or the flowers, species of the birds or dogs, and type of the aircrafts. These differences are what make FGVC a challenging problem, as there are significant intra-class differences among the sub-categories, and at the same time, items from different classes may look similar. In contrast with regular object classification techniques, FGVC aims to solve the identification of particular subcategories from a given category \cite{ge2016fine,hu2019see}. 

Convolutional Neural Networks (CNNs) have been extensively used for various applications in computer vision. To achieve good performance with CNNs, typically we need large amounts of labeled data. However, it is a tedious process to collect labeled fine-grained datasets. That is why there are not many FGVC datasets, and existing datasets are not as large compared to standard image recognition datasets like ImageNet \cite{deng2009imagenet}. Normally, a model pre-trained on large scale datasets such as ImageNet is used, and that model is then fine-tuned using data from an FGVC dataset. Typically, FGVC datasets are not too big, so it becomes critical to design methods that can compensate for the limited amount of data. In this paper, we investigate some techniques that allow the model to learn features more effectively, and that perform well on large scale datasets with fine-grained categories.

Generally, there are two domains involved in fine-tuning a network. One is the source domain, which typically includes large scale image datasets like ImageNet \cite{deng2009imagenet}, where initial models are pre-trained. Another is the target domain, where data is used to fine-tune the pre-trained models. In this paper, the target domain is FGVC datasets, and we are interested in developing techniques that can boost accuracy on these type of datasets. Modern FGVC methods use pre-trained networks with ImageNet dataset to a large extent. We explore the possibility of achieving better accuracy than what has been achieved so far using ImageNet. A model first learns useful features from a large amount of training data, and is then fine-tuned on a more evenly-distributed subset to balance the efforts of the network among different categories and transfer the already learned features. 

In short, our research tries to address two questions: 
1) What approaches beyond transfer learning do we need to take to boost the performance on FGVC datasets?
2) How can we determine which large scale source domain we choose, given that the target domain is FGVC?

 We calculate the domain similarity score between the source and target domains. This score gives us a clear picture of selecting the source domain for transfer learning to achieve better accuracy in the target domain. Then, we focus on a visual attention guided network for data augmentation. As FGVC datasets are relatively smaller in size, we leverage the feature learning from fine-tuning as well as data augmentation to achieve better accuracy. The performance of the combination of these two strategies outperforms the baseline approach. 

In summary, the main contributions of this work are:
\begin{enumerate}
    \item We propose a simple yet effective improvement over the recently proposed Weakly Supervised Data Augmentation Network (WS-DAN) \cite{hu2019see}, which is used for generating attention maps to extract sequential local features to tackle the FGVC challenge. A domain similarity score can play a vital role before applying transfer learning. Based on the score, we decide which source domain is necessary to use for transfer learning. Then, we can employ WS-DAN \cite{hu2019see} to achieve  better results among FGVC datasets.   
    \item We demonstrate a domain adaptive transfer learning approach, that combines with visual attention based data augmentation, and that can achieve state-of-the-art results on CUB200-2011 \cite{wah2011caltech}, and Flowers-102 \cite{flowers}, and  FGVC-Aircrafts \cite{maji2013fine} datasets. Additionally, we match the current state-of-the-art accuracy on Stanford Cars \cite{krause20133d}, Stanford Dogs  \cite{khosla2011novel} datasets. 
    \item We present the relationship of top-1 accuracy and domain score on six commonly used FGVC datasets. We illustrate the effect of image resolution in transfer learning in detail. 
\end{enumerate}

\section{Related Work}
In this section, we present a brief overview of data augmentation, fine-grained visual categorization, visual attention mechanism and transfer learning. 

%-------------------------------------------------------------------------
\subsection{Data Augmentation}
 Machine learning theory suggests that a model can be more generalized and robust if it has been trained on a dataset with higher diversity. However, it is a very difficult and time-consuming task to collect and label all the images which involve these variations \cite{zhang2019survey}. Data augmentation methods are proposed to address this issue by adding the amount and diversity of training samples. Various methods have been proposed focusing on random spatial image augmentation, specifically involving in rotation variation, scale variation, translation, and deformation, etc. \cite{hu2019see}.
Classical augmentation methods are widely adopted in deep learning techniques. 

The main drawback of random data augmentation is low model accuracy. Additionally, it suffers from generating a lot of unavoidable noisy data. Various methods have been proposed to consider data distribution rather than random data augmentation. A search space based data augmentation method has been proposed \cite{cubuk2018autoaugment}. It can automatically search for improving data augmentation policies in order to obtain better validation accuracy. In contrast, we leverage WS-DAN \cite{hu2019see}, which generates augmented data from visual attention features of the image. Peng \textit{et al.} proposed a method for human pose estimation, by introducing an augmentation network whose task is to generate hard data online, thus improving the robustness of models \cite{peng2018jointly}. Nevertheless, their augmentation system is complicated and less accurate compared to the network that we experimented with. Additionally, attention-aware data segmentation is more simple and proven effective in terms of accuracy. 

\subsection{Fine-Grained Visual Categorization}

Fine-grained Visual Categorization (FGVC) is a challenging problem in the field of computer vision. Normally, object classification is used for categorize different objects in the image, such as humans, animals, cars, trees, etc. In contrast, fine-grained image classification concentrates more on detecting sub-categories of a given category, like various types of birds, dogs or cars. % give examples
The purpose of FGVC is to find subtle differences among various categories of a dataset. It presents significant  challenges for building a model that generalizes patterns. FGVC is useful in a wide range of applications such as image captioning \cite{anne2016deep}, image generation \cite{bao2017cvae}, image search engines, and so on.

Various methods have been developed to differentiate fine-grained categories. Due to the remarkable success of deep learning, most of the recognition works depend on the powerful convolutional deep features. Several methods were proposed to solve large scale real problems \cite{simon2015neural,he2016deep,szegedy2016rethinking}.  However, it is relatively hard for the basic models to focus on very precise differences of an object's parts without adding special modules \cite{hu2019see}. A weakly supervised learning-based approach was adapted to generate class-specific location maps by using pooling methods \cite{lin2015bilinear}. Adversarial Complementary Learning (ACoL) \cite{zhang2018adversarial} is a weakly supervised approach to identify entire objects by training two adversarial complementary classifiers, which aims at locating several parts of objects and detects complementary regions of the same object. 
However, their method fails to accurately locate the parts of the objects due to having only two complementary regions.
On the contrary, our proposed approach depends on attention-guided data augmentation and domain adaptive transfer learning. Our method extracts fine-grained discriminative features and provides a generalization of domain features to achieve state-of-the-art performance in terms of accuracy.  

\subsection{Attention}
Attention mechanisms have been getting a lot of popularity in the deep learning area. Visual attention has been already used for FGVC. Xiao \textit{et al.} proposed a two-way attention method (object-level attention and part-level attention) to train domain-specific deep networks \cite{xiao2015application}. Fu \textit{et al.} proposed an approach that can predict the location of one attention area and extract corresponding features \cite{fu2017look}. However, this method can only focus on a local object's parts at the same time. Zheng \textit{et al.} addressed this issue and introduced Multi-Attention CNN (MA-CNN) \cite{zheng2017learning}, which can simultaneously focus on multiple body parts. However, selected parts of the object are limited and the number of selected parts is fixed (2 or 4), which might hamper accuracy. 

The works mentioned above mostly focus on object localization. In contrast, our research concentrates more on data augmentation with visual attention, which has not been much explored. We use the attention mechanism for data augmentation purposes. Moreover, the benefit of guided attention based data augmentation \cite{hu2019see} helps the network to locate object precisely, which helps our trained model learn about closer object details and hence, improve the predictions.  

%-------------------------------------------------------------------------
\subsection{Transfer Learning}
The purpose of transfer learning is to improve the performance of a learning algorithm by utilizing knowledge that is acquired from previously solved similar problems. CNNs have been widely used for transfer learning. They are mostly used in the form of pre-trained networks that serve as feature extractors \cite{sharif2014cnn,donahue2014decaf}.

Considerable amounts of effort have been made to understand transfer learning \cite{yosinski2014transferable,sun2017revisiting,azizpour2015factors}. 
Initial weights for a certain network can be obtained from an already-trained network even if the network is used for different tasks  \cite{yosinski2014transferable}. Some prior work has shown some results on transfer learning and domain similarity \cite{cui2018large}. Their contribution mostly addresses the effect of image resolution on large scale datasets and choosing different subsets of datasets to boost accuracy. In our work, we show that domain adaptive transfer learning can be  useful if we also incorporate visual attention based data augmentation. 

Unlike previous works, our proposed technique takes account of domain adaptive transfer learning  between the source and target domains. Then, it incorporates the attention-driven approach for data augmentation. Our main goal is to guide the training model to learn relevant features from the source domain and augment data with the visual attention of the target domain. The combination of two processes can be useful to achieve better performance.

\section{Domain Adaptive Transfer Learning (DATL)}
In our research, we explore the way of determining similarity between the source and target domains. Additionally, we describe the attention aware data augmentation technique, WS-DAN in detail. We consider different types of large scale datasets to find out the similarity score between large scale datasets and FGVC datasets. Then, we compute  domain similarity score firstly. Based on the domain similarity score we choose large scale datasets for transfer learning and then we perform WS-DAN to evaluate the accuracy. 

\subsection{Domain Similarity}
Generally, transfer learning performs better if it has been trained on bigger datasets. Chen \textit{et al.} showed that transfer learning performance increases logarithmically with the number of data \cite{sun2017revisiting}. In our work, we observe that using a bigger dataset does not always provide a more accurate result. Yosinski \textit{et al.} \cite{yosinski2014transferable} mentions that there is some correlation between the transferability of a network from the source task to the target task and the distance between the source and target tasks. Furthermore, they show fine-tuning on a pre-trained network towards a target task can boost performance. Our domain adaptive transfer learning approach is inspired from Cui \textit{et al.} \cite{cui2018large} who introduce a method which can calculate domain similarity by the Earth Mover's Distance (EMD) \cite{rubner2000earth}. Furthermore, they show transfer learning can be treated as moving image sets from the source domain $S$ to the target domain $T$. The domain similarity \cite{cui2018large} can be defined
\begin{equation}
    d(S,T)=EMD(S,T) = \frac{\sum_{i=1,j=1}^{m,n} f_{i,j}d_{i,j}}{\sum_{i=1,j=1}^{m,n} f_{i,j}}
\end{equation}
where $s_i$ is $i$-th category in $S$ and $t_j$ is $j$-th in $T$, $d_{i,j}= ||g(s_i) - g(t_j)||$ , feature extractor $g(.)$ of an image and the optimal flow $f_{i,j}$ computes total work as a EMD minimization problem. Finally, the similarity is calculated as: 
\begin{equation}
    sim(S,T) = e^{-\gamma d(S,T)}
\end{equation}

where $\gamma$ is a regularization constant of value 0.01. 

Domain similarly score can be calculated between the source and target domain. In our approach, we use large scale datasets as source domains, and target domains are selected from six commonly used FGVC datasets. After calculating the similarity score, we choose top k categories with the highest domain similarity. 

\subsection{Attention Aware Data Augmentation} 
In our method, we consider using the Weakly Supervised Data Augmentation Network (WS-DAN) \cite{hu2019see}. Firstly, we extract features of the image I  and feature maps $F \in R^{H \times W \times C}$, where H, W, and C correspond to height, width, and number of channels of a feature layer. Then, we generate attention maps $A \in R^{H \times W \times M}$ from feature maps, where M is the number of attention maps. One more critical component is bi-linear attention pooling, which is used to extract features from part objects. Element-wise multiplication between feature maps and attention maps is computed to get part-feature maps, and then, pooling operation is applied on part-feature maps afterward. Randomly generated data from augmentation is not much efficient. However, attention maps can be handy for data augmentation. This way model can be guided to focus on essential parts of the data and augment those data to the network. With an augmentation map, part's region can be zoomed, and detailed features can be extracted. This process is called attention cropping. Attention maps can represent similar object's part. Attention dropping can be applied to the network to distinguish multiple object's part. Both attention cropping and attention dropping are controlled through a threshold value.

During the training process, no bounding box or keypoints based annotation is available. For each particular training image, attention maps are generated to represent the distinguishable part of object. Attention, guided data augmentation component, is responsible for selecting attention maps efficiently utilizing attention cropping and attention dropping. Bilinear Attention Pooling (BAP) is used to extract features from the object's parts. Element-wise multiplication between the feature maps and attention map are used to generate a part feature matrix. In the last step, the original data, along with attention generated augmented data, are trained as input data.

During the testing process, in the beginning, the object's categories probability and attention maps are produced from input images. Then, the selected part of the object can be enlarged to refine the category's probability. The final prediction is evaluated as the average of those two probabilities. The process of final prediction \cite{hu2019see} is presented as Algorithm 1. 

\begin{algorithm}[tb]
\caption{Attention Aware Fine-grained Categorization }
\label{alg:algorithm}
\textbf{Input}: Trained model with WS-DAN and Raw Image I\\
\textbf{Output}: Classification Accuracy
\begin{algorithmic}[1] %[1] enables line numbers
\Statex
\State Calculate coarse-grained probability $p_1: p_1 = W(I)$ and generate attention maps A
\Statex
\State Calculate object map $A_m$ from A and obtain bounding box $B$ from $A_m$
\Statex
\State Zoom in the region $B$ as $I_b$ 
\Statex
\State Predict fine-grained probability $p_2: p_2 = W(I_b)$ 
\Statex
\State Calculate final probability $p = (0.5)*(p_1 + p_2)$ 
\Statex
\State \textbf{return} p
\end{algorithmic}
\end{algorithm}

The training process is illustrated in Figure 1. During training process, no bounding box or keypoints based annotation are available. For each particular training image, attention maps are generated to represent the distinguishable part of object. Attention guided data augmentation component is responsible to select attention maps efficiently utilizing attention cropping and attention dropping. Bilinear Attention Pooling (BAP) is used to extract feature from object's parts. Element-wise multiplication between the feature maps and attention map is used to generate part feature matrix. In the last step, the original data along with attention generated augmented data are trained as input data. 

Figure 2 shows the illustration of testing process. Firstly, the object's categories probability and attention maps are produced from input images. Then, selected part of the object can be enlarged to refine the categories probability. The final prediction is evaluated as the average of those two probabilities.

\begin{figure*}[h]

%   %\includegraphics[width=0.8\linewidth]{egfigure.eps}

\includegraphics[width=\textwidth,height=6cm]{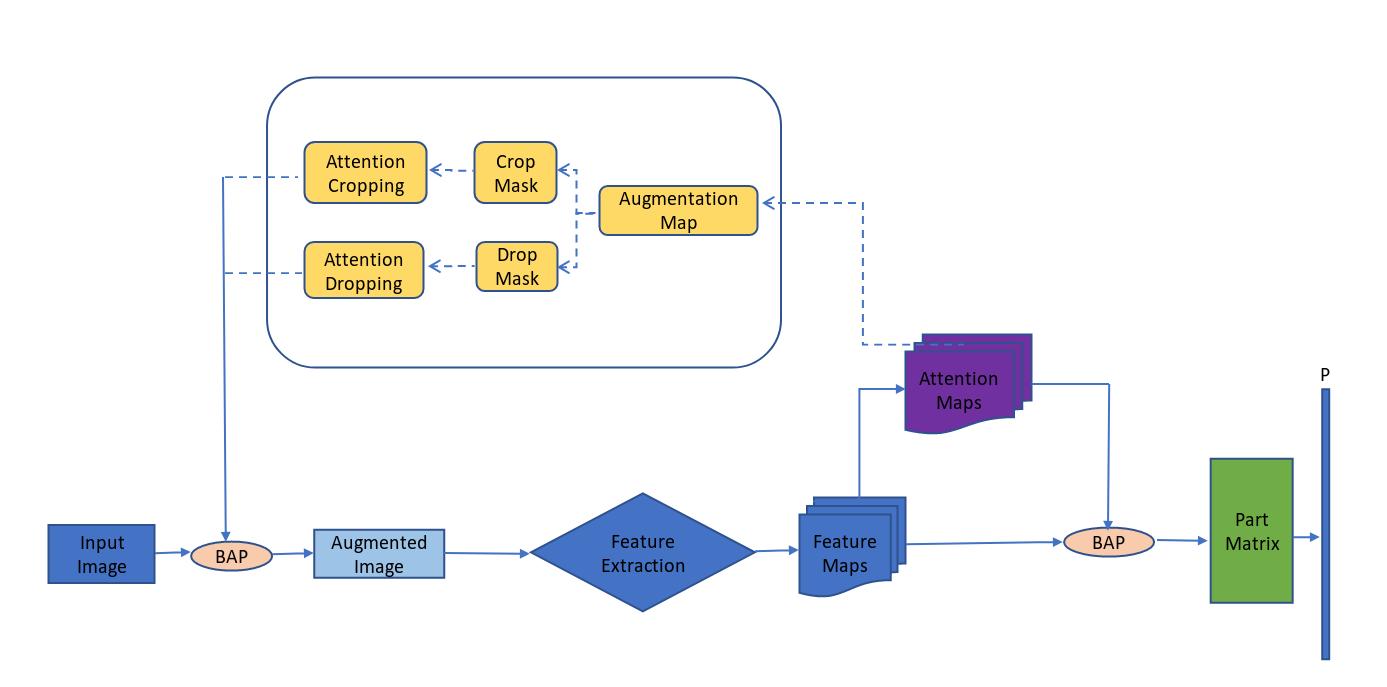}

\caption{Weakly Supervised Data Augmentation Network  \cite{hu2019see} Training Process.}

\end{figure*}

\begin{figure*}[h]

%   %\includegraphics[width=0.8\linewidth]{egfigure.eps}

\includegraphics[width=\textwidth,height=6cm]{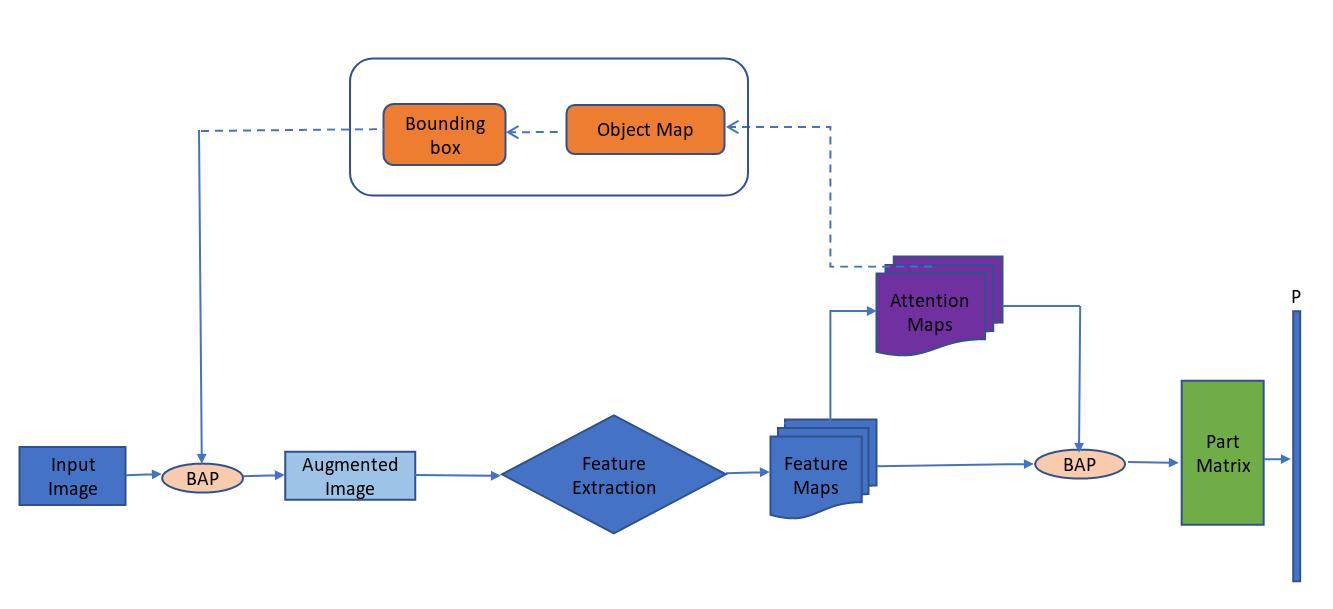}

\caption{Weakly Supervised Data Augmentation Network  \cite{hu2019see} Testing Process.}

\end{figure*}

\subsection{Visualization of Augmented Data}
We visualize the attention-guided data augmentation in CUB200-2011, Food-101, Flowers-102, Stanford Car, Stanford Dog and FGVC-Aircraft respectively in Figure 3-8. 

\graphicspath{ {./images/} }

\begin{figure}[h]

%   %\includegraphics[width=0.8\linewidth]{egfigure.eps}
\centering
\includegraphics[width=8cm,height=2cm]{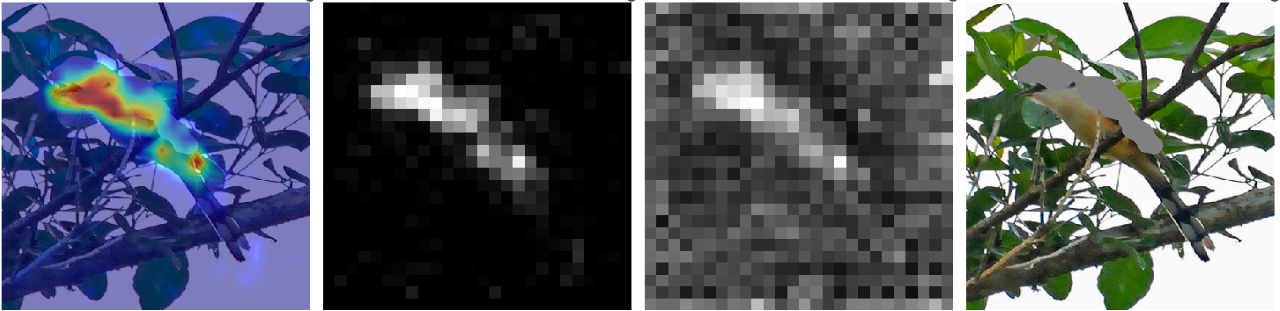} \\
\includegraphics[width=8cm,height=2cm]{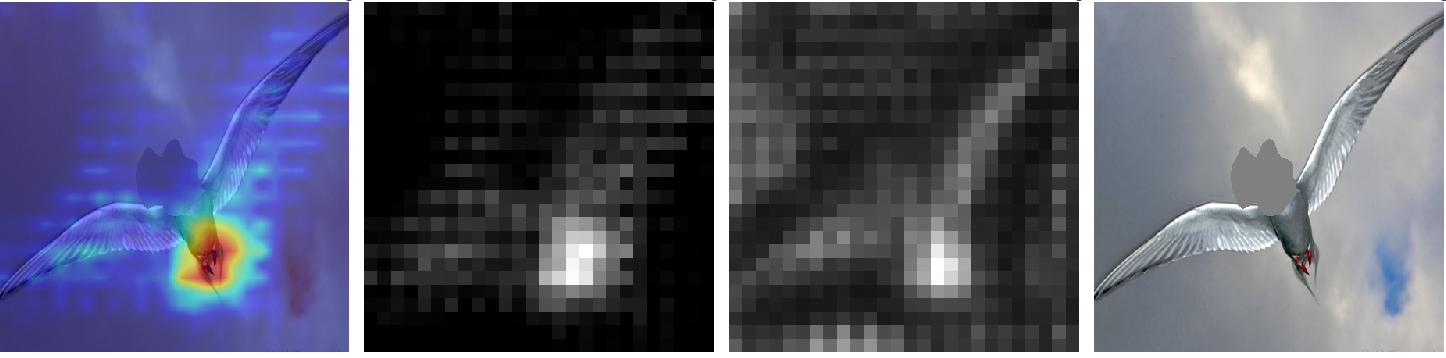}

\caption{Visual attention on image, Attention Maps, Feature Maps, Attention Dropping on CUB200-2011 dataset (left to right respectively) \cite{hu2019see}.}
\label{fig:short}

\end{figure}

\begin{figure}[h]

%   %\includegraphics[width=0.8\linewidth]{egfigure.eps}
\centering
\includegraphics[width=8cm,height=2cm]{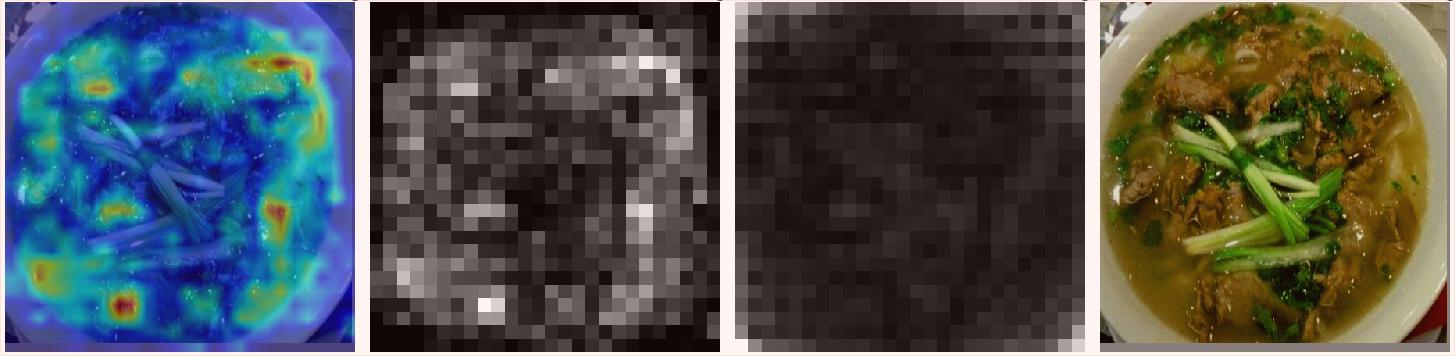} \\
\includegraphics[width=8cm,height=2cm]{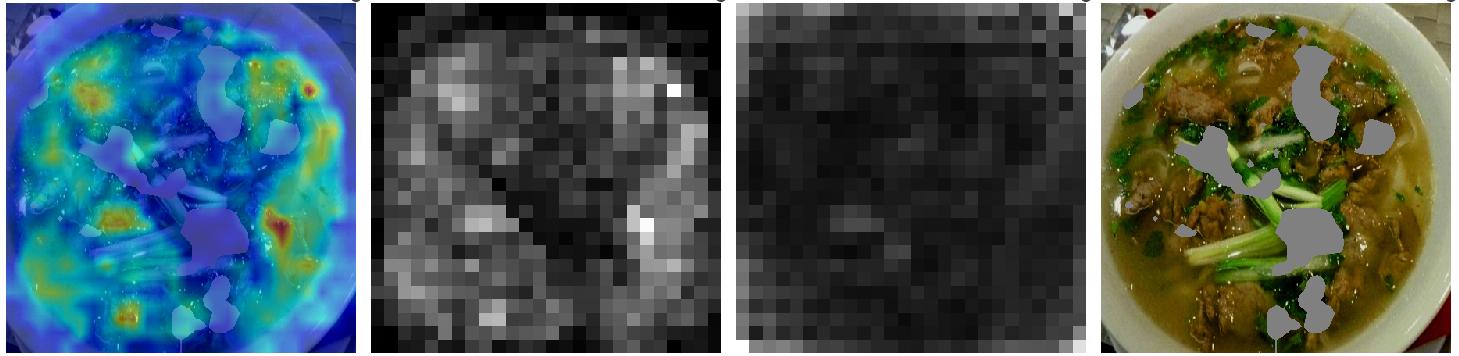}

\caption{Visual attention on image, Attention Maps, Feature Maps, Attention Dropping on Food-101 dataset (left to right respectively) \cite{hu2019see}.}
\label{fig:short}

\end{figure}

\begin{figure}[h]

%   %\includegraphics[width=0.8\linewidth]{egfigure.eps}
\centering
\includegraphics[width=8cm,height=2cm]{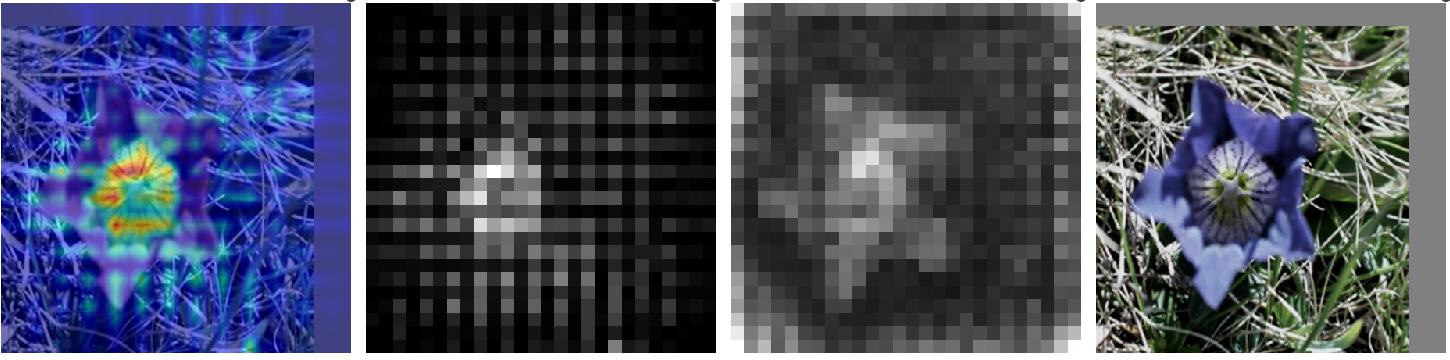} \\
\includegraphics[width=8cm,height=2cm]{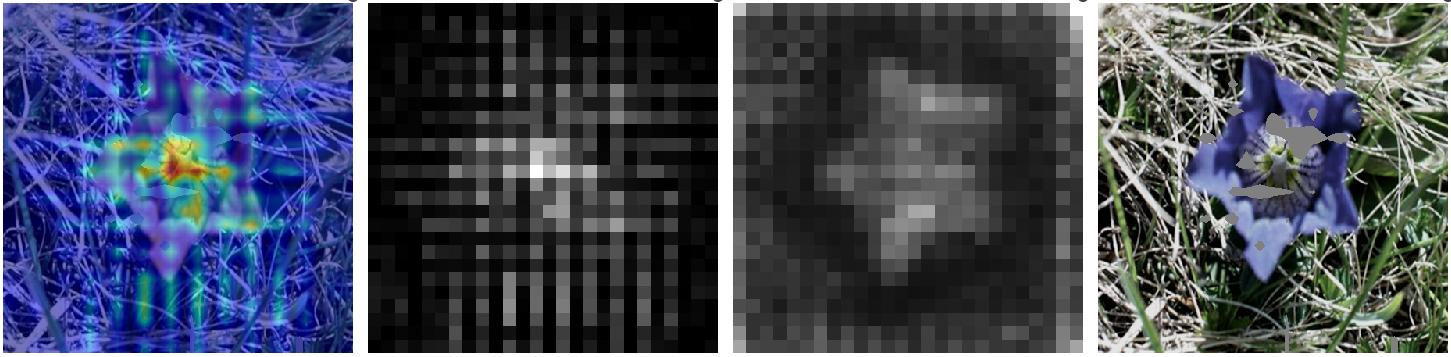}

\caption{Visual attention on image, Attention Maps, Feature Maps, Attention Dropping on Flowers-102 dataset (left to right respectively) \cite{hu2019see}.}
\label{fig:short}

\end{figure}

\begin{figure}[h]

%   %\includegraphics[width=0.8\linewidth]{egfigure.eps}
\centering
\includegraphics[width=8cm,height=2cm]{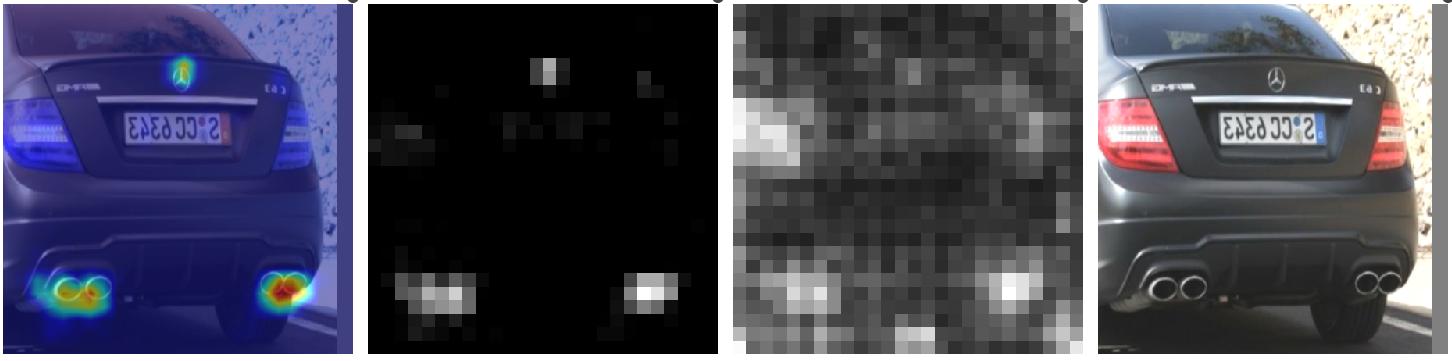} \\
\includegraphics[width=8cm,height=2cm]{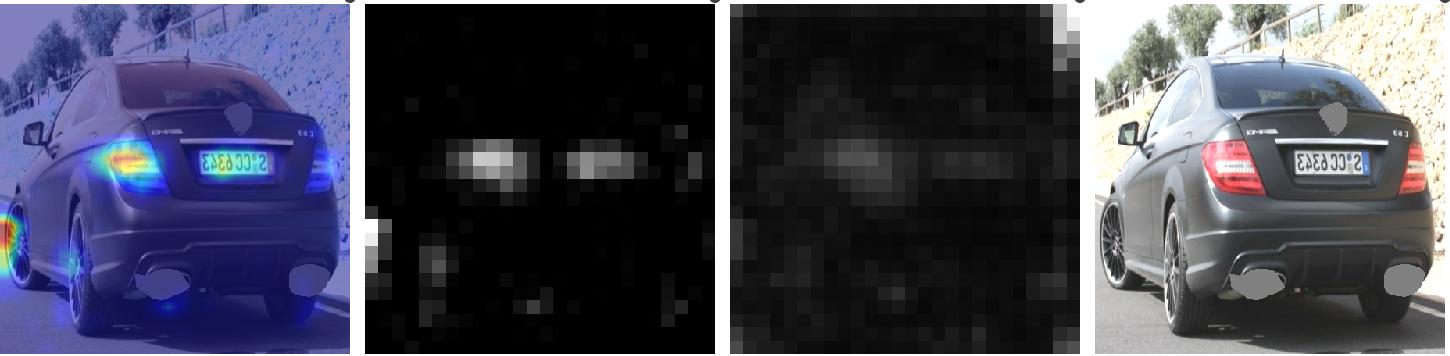}

\caption{Visual attention on image, Attention Maps, Feature Maps, Attention Dropping on Stanford Car dataset (left to right respectively) \cite{hu2019see}.}
\label{fig:short}

\end{figure}

\begin{figure}[h]

%   %\includegraphics[width=0.8\linewidth]{egfigure.eps}
\centering
\includegraphics[width=8cm,height=1.5cm]{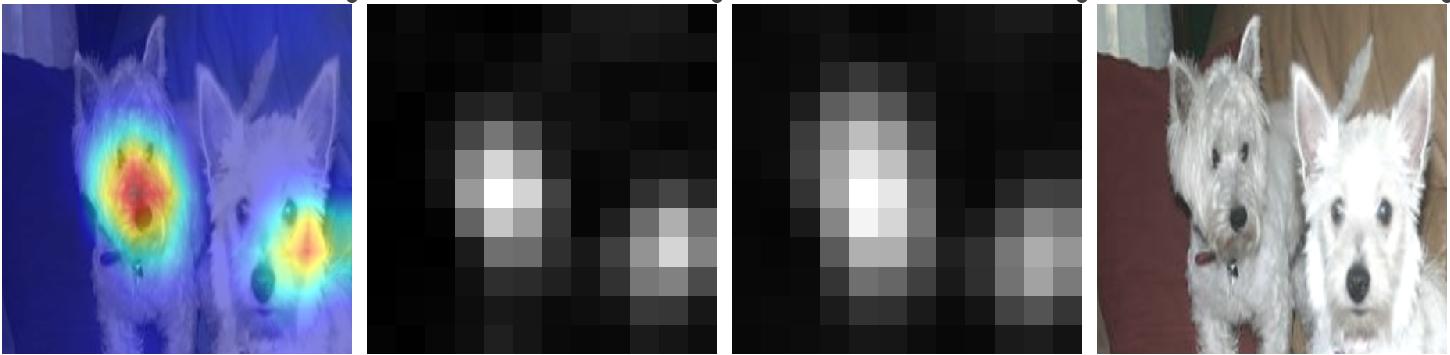} \\
\includegraphics[width=8cm,height=1.5cm]{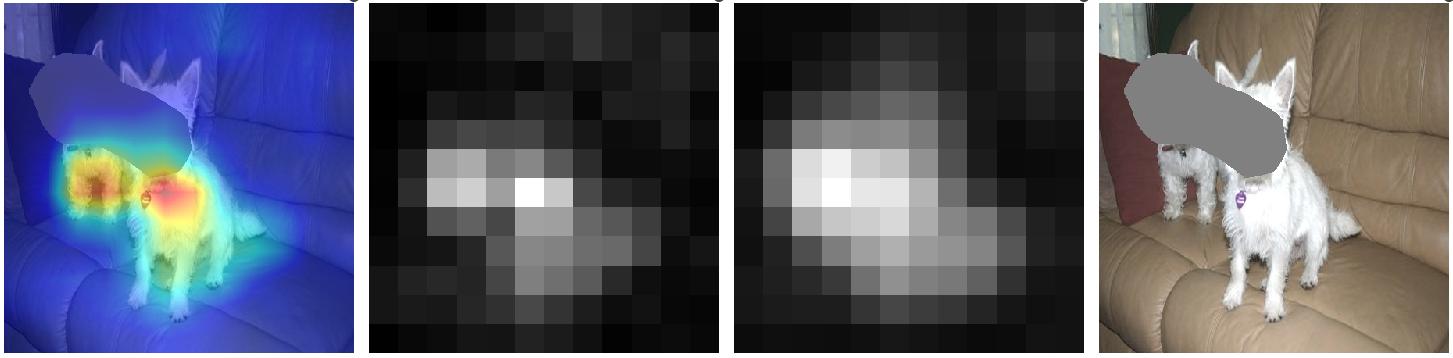}

\caption{Visual attention on image, Attention Maps, Feature Maps, Attention Dropping on Stanford Dog dataset (left to right respectively) \cite{hu2019see}.}
\label{fig:short}

\end{figure}

\begin{figure}[h]

%   %\includegraphics[width=0.8\linewidth]{egfigure.eps}
\centering
\includegraphics[width=8cm,height=1.5cm]{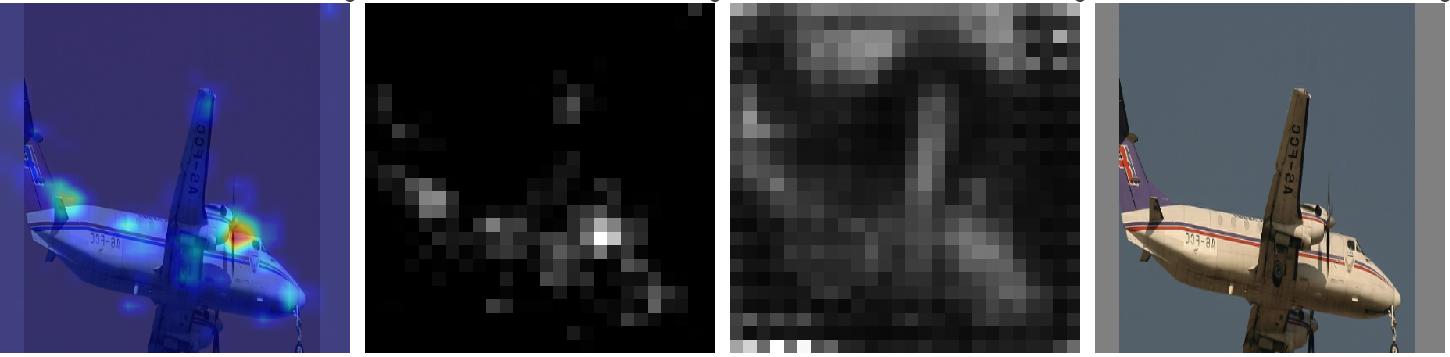} \\
\includegraphics[width=8cm,height=1.5cm]{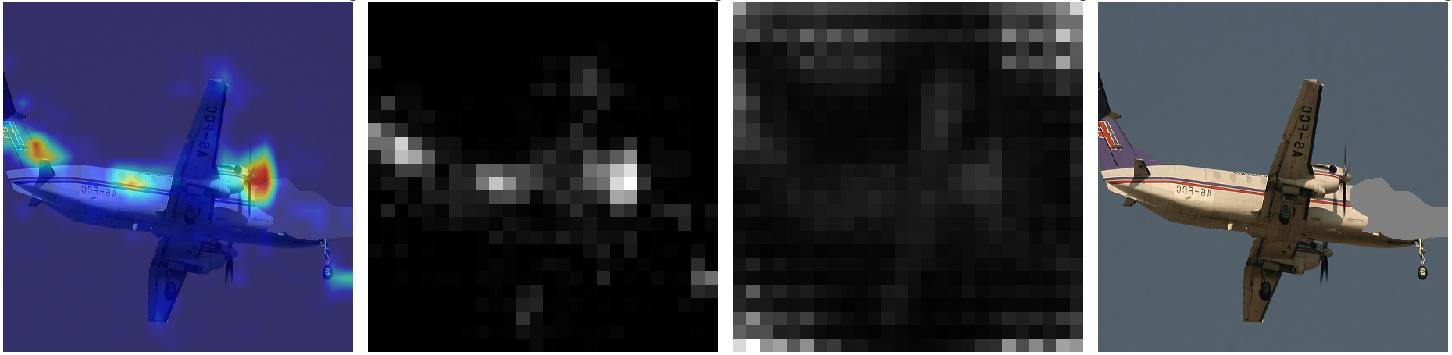}

\caption{Visual attention on image, Attention Maps, Feature Maps, Attention Dropping on FGVC-Aircraft dataset (left to right respectively) \cite{hu2019see}.}
\label{fig:short}

\end{figure}

\subsection{Loss Function}
The loss function of the network is derived from center loss \cite{wen2016centerloss}, which has been proposed to tackle face recognition issues. Here, we adopt attention regularization loss \cite{hu2019see} for the attention learning process. The idea is to minimize the intra-class variations while keeping the features of inter-class features differentiable. So, penalizing the features variation that belong to same part of object which is important for fine-grained category. The loss function can be defined as:
\begin{equation}
    L_A = \sum_{k=1}^M ||f_k - c_k||_2^2
\end{equation}
where $M$ is number of attention maps, $f_k$ is the part feature and $c_k$ is its part's feature center of $k$th object. $c_k$ can be updated by moving average and initialized as zero, and the update rate is $\beta$ . 
\begin{equation}
    c_k \leftarrow c_k + \beta(f_k - c_k)
\end{equation}

\section{Experiments}
In this section, we show comprehensive experiments to
verify the effectiveness of our approach. Firstly, we calculate the domain similarity score using EMD \cite{rubner2000earth} to demonstrate the relationship between the source and target domains. Then we compare our model with the state-of-the-art methods on six publicly available fine-grained visual categorization datasets. Following this, we perform additional experiments to demonstrate the effect of image resolution on transfer learning. We compare input images in the iNaturalist dataset from $299 \times 299$ to $448 \times 448$ to observe the effect in terms of accuracy. We have trained the baseline inceptionV3 model with iNaturalist datasets.
Additionally, we combine both iNaturalist and imageNet dataset to make a bigger dataset. We perform detailed experimental studies with different types of large scale datasets and apply the WS-DAN method to observe the impact. The training loss curve and top-1 accuracy curve are presented in Figures 9 and 10, respectively.

\begin{figure}[t]

%   %\includegraphics[width=0.8\linewidth]{egfigure.eps}
\centering
\includegraphics[width=10cm,height=6cm]{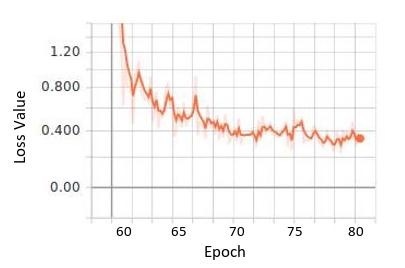}

\caption{Training loss on CUB200-2011 dataset} %\cite{hu2019see}}
\label{fig:short}

\end{figure}

\begin{figure}[t]

%   %\includegraphics[width=0.8\linewidth]{egfigure.eps}
\centering
\includegraphics[width=10cm,height=6cm]{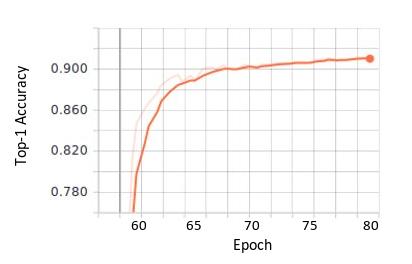}

\caption{Accuracy on CUB200-2011 dataset}%\cite{hu2019see}}
\label{fig:short}

\end{figure}

\subsection{Datasets}

We present a detailed overview of the datasets that we use for our experiments.

\textbf{ImageNet}: The ImageNet \cite{deng2009imagenet} contains 1.28 million training images and 50 thousand validation images along with 1,000 categories. 

\textbf{iNaturalist(iNat)} : The iNat dataset, introduced in 2017 \cite{van2018inaturalist}, contains more than 665,000 training and around 10000 test images from more than 5000 natural fine-grained categories. Those categories include different types of mammals, birds, insects, plants, and more.  This dataset is quite imbalanced and varies a lot in terms of the number of images per category.

\textbf{Fine-grained object classification datasets}: Table 1 summarizes the information of each dataset in detail.

%------------------------------------------------------------------------

\subsection{Implementation Details}
In our experiment, we used Tensorflow \cite{abadi2016tensorflow} to train all the models on multiple Nvidia Geforce GTX 1080Ti GPUs. The machine has Intel Core-i7-5930k CPU@ 3.50GHz x 12 processors with 64GB of memory. During training, we adopted Inception v3 \cite{szegedy2016rethinking} as the backbone network. We employed WS-DAN \cite{hu2019see} technique to perform experiments to demonstrate the effectiveness of transfer learning.  For all the datasets, we used Stochastic Gradient Descent (SGD) with a momentum of 0.9, the number of epoch 80, mini-batch size 12. The initial learning rate was set to 0.001, with exponential decay of 0.8 after every 2 epochs. 

\begin{figure}[h]
\centering
\includegraphics[width=\linewidth,height=6cm]{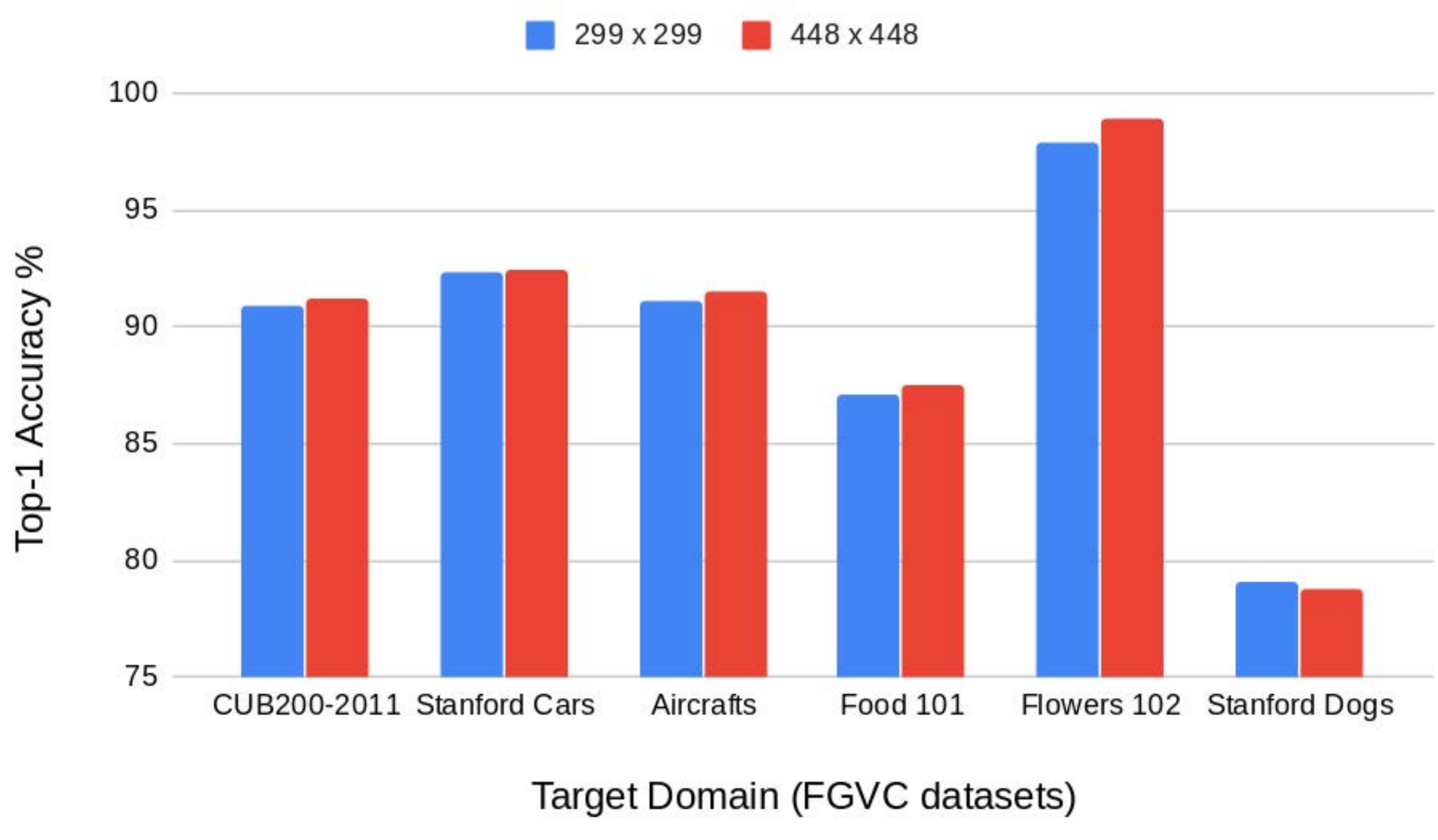} % first figure itself
\caption{Effect of transfer learning with different sizes of image resolution on iNat dataset.}
\end{figure}
\begin{figure}
\centering
\centering
\includegraphics[width=\linewidth,height=6cm]{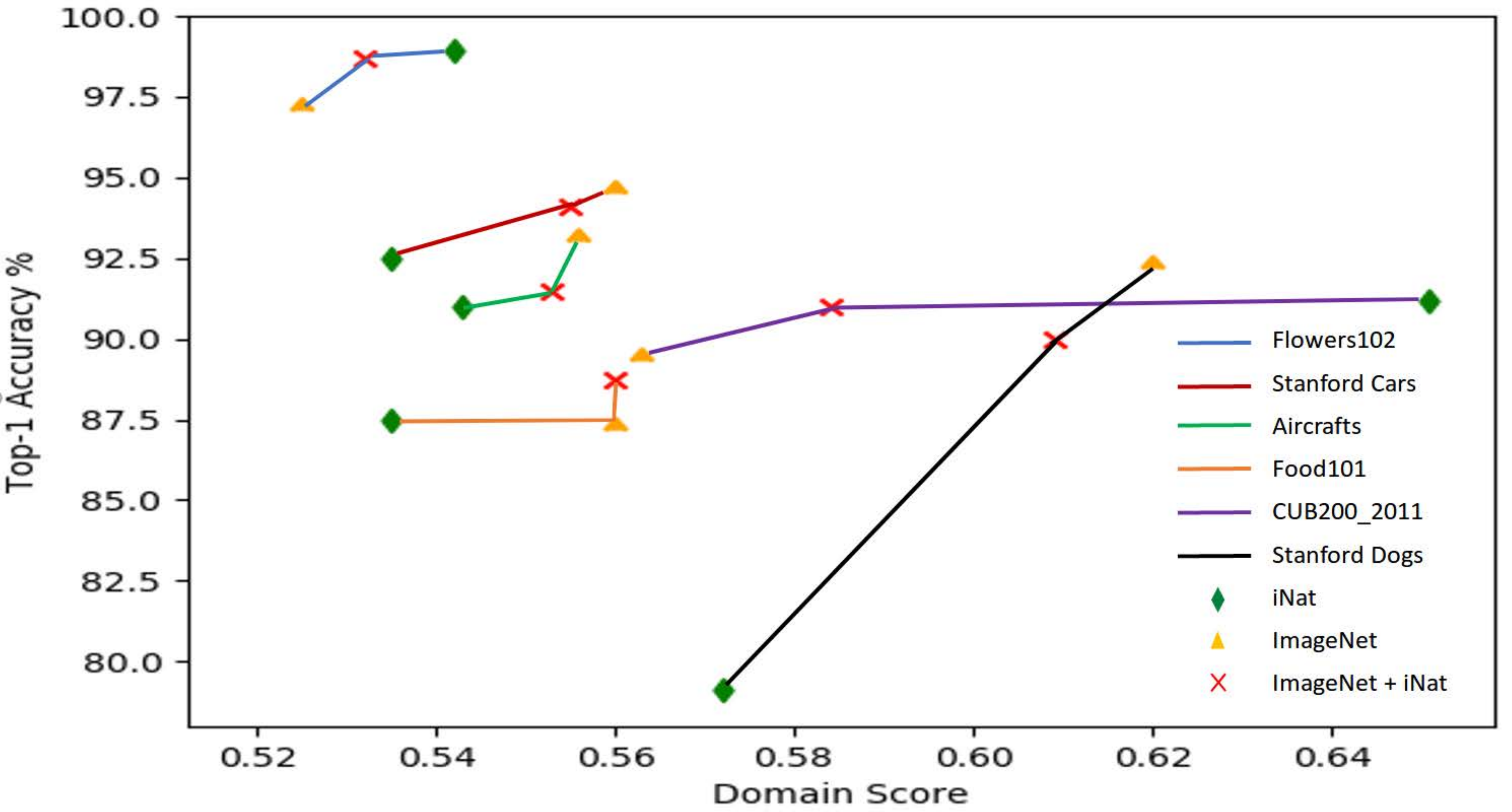} % second figure itself
\caption{Correlation between transfer learning accuracy and domain similarity score between the source and target domain. Each colored line represents a target domain.}
\end{figure}

\begin{table}[t]
\centering
\begin{center}
\caption{Six commonly used FGVC datasets.}
\begin{tabular}{|c|c|c|c|c|}
\hline
Datasets & Objects & Classes & Training & Test \\
\hline
CUB200-2011 & Bird & 200 & 5,994 & 5,794 \\
FGVC-Aircraft & Aircraft & 100 & 6,667 & 3,333 \\
Stanford Cars & Car & 196 & 8,144 & 8,041 \\
Stanford Dogs & Dog & 120 & 12,000 & 8,580 \\
Flowers-102 & Flowers & 102 & 2,040 & 6,149 \\
Food-101 & Food & 101 & 75,750 & 25,250 \\
%NABirds & Birds & 555 & 23,929 & 24,633 \\
\hline
\end{tabular}
\end{center}
\end{table}

\begin{table}[h]
\centering
\begin{center}
\begin{tabular}{|c|c|c|c|c|c|c|}
\hline
Source Domain & CUB200-2011 & Stanford Cars & Aircrafts & Food101 & Flowers102 & Stanford  Dogs \\
\hline
ImageNet & 0.563 & 0.56 & 0.556 & 0.56 & 0.525 & 0.619 \\
iNaturalist  & 0.651 & 0.535 & 0.543 & 0.535 & 0.542 & 0.572 \\
ImageNet + iNat & 0.584 & 0.555 & 0.553 & 0.56 & 0.532 & 0.608 \\
\hline
\end{tabular}
\end{center}
\caption{Comparison on domain similarity score between source datasets and target datasets}
\end{table}

\begin{table}[h]
\centering
\begin{center}
\caption{Comparison to different types of FGVC datasets. Each row represents a network pre-trained on source domain for transfer learning and each column represents top-1 image classification accuracy by fine-tuning on the target domain.}
\begin{tabular}{|c|c|c|c|c|c|c|}
\hline
\specialcell{Method} &  \specialcell{CUB200 \\ 2011} & \specialcell{Stanford \\ Cars} & Aircrafts & \specialcell{Food \\ 101} & \specialcell{Flowers \\ 102} & \specialcell{Stanford \\ Dogs} \\
\hline
ImageNet & 82.8 & 91.3 & 85.5 & 88.6 & 96.2 & 84.2 \\
\hline
\specialcell{ImageNet on WS-DAN} & 89.3 & \textbf{94.5} & \textbf{93.0} & 87.2 & 97.1 & \textbf{92.2} \\
\hline
\specialcell{iNat on WS-DAN} & \textbf{91.2} & 92.5 & 91.0 & 87.5 & \textbf{98.9} & 79.1 \\
\hline
\specialcell{ImageNet + iNat on WS-DAN }  & 91.0 & 94.1 & 91.5 & \textbf{88.7} & 98.7 & 90.0  \\

\hline
\end{tabular}
\end{center}
\end{table}

\section{Results}
When training a CNN, input images are often preprocessed to match a specific size. Higher resolution images usually contain essential information and precise details that are important to visual recognition. We compare results on six FGVC datasets with different sizes of image resolution of the iNat dataset. In summary, images with higher resolution yields better accuracy except for the Stanford Dogs dataset. Figure 11 represents the effect of transfer learning with various sizes of image resolution on iNat dataset. 

\begin{table}[h]
\centering
%\begin{center}
\caption{Comparison in terms of accuracy with existing FGVC methods.}
\begin{tabular}{|c|c|c|c|c|c|c|}
\hline
\specialcell{Method} &  \specialcell{CUB200 \\ 2011} & \specialcell{Stanford \\ Cars} & Aircrafts & \specialcell{Food \\ 101} & \specialcell{Flowers \\ 102} & \specialcell{Stanford \\ Dogs} \\
\hline
Bilinear-CNN \cite{lin2015bilinear} & 84.1 & 91.3 & 84.1 & 82.4 & - & - \\
DLA \cite{yu2018deep} & 85.1 & 94.1 & 92.6 & 89.7 & - & - \\
RA-CNN \cite{fu2017look} & 85.4 & 92.5 & - & - & - & 87.3 \\
\specialcell{Improved Bilinear-CNN \cite{lin2017improved}} & 85.8 & 92.0 & 88.5 & - & - & - \\
GP-256 \cite{zheng2017learning} & 85.8 & 92.8 & 89.8 & - & - & - \\
MA-CNN \cite{fu2017look} & 86.5 & 92.8 & 89.9 & - & - & - \\
DFL-CNN \cite{wang2018learning} & 87.4 & 93.8 & 92.0 & - & - & - \\
MPN-COV \cite{li2018towards} & 88.7 & 93.3 & 91.4 & - & - & - \\
Subset B \cite{cui2018large} & 89.6 & 93.5 & 90.7 & 90.4 & - & 88.0 \\
WS-DAN \cite{hu2019see} & 89.4 & 94.5 & 93.0 & 87.2 & 97.1 & 92.2 \\
\textbf{DATL + WS-DAN} & \textbf{91.2} & 94.5 & \textbf{93.1} & 88.7 & \textbf{98.9} & \textbf{92.2} \\

\hline
\end{tabular}
%\end{center}

\end{table}

In Table 3, we present the top-1 accuracy of the target domains on various source domains. These results show the impact of transfer learning from a pre-trained model. Large scale datasets are essential for getting improved accuracy when transfer learning is conducted. ImageNet dataset is much larger than iNat dataset; still, it shows worse accuracy in the CUB200-2011 dataset. So, we cannot conclude that using a bigger dataset with transfer learning can always yield better results. Moreover, the domain similarity score also supports this hypothesis. Hence, transfer learning can be effective if the target domain can be trained with similar source domain. 

 We compare our method with state-of-the-art baselines on six commonly used fine-grained categorization datasets. The summary of the comparison is presented in Table 4. In Table 2, we show the domain similarity score between the source and various target domains. We visually represent the relationship between the top-1 accuracy and the domain similarity score. We can observe from Figure 12 that the domain similarity score positively correlated with transfer learning accuracy between large scale datasets and FGVC datasets. Each marker represents a source domain. With the right selection of source domain, better transfer learning performance can be achieved. For example, the domain similarity score between iNat and CUB200-2011 is around \textbf{0.65}, which is the reason it shows higher accuracy \textbf{(91.2)} when iNat is used as pre-training the source domain compared to others. For Flowers-102 dataset, the accuracy is \textbf{98.9} with iNat as the source domain which has the highest domain simiarity score \textbf{0.54}, among other source domains. Similarly, Stanford Cars, Stanford Dogs and Aircrafts dataset show higher domain similarity score supports better accuracy.  Only for the Food101 dataset, the accuracy from transfer learning remains similar while domain similarity changes. We believe this is due to having a large number of training images in Food101. Consequently, the target domain contains enough data and transfer learning is not as useful. We can observe that both ImageNet and iNat are highly biased, achieving dramatically different transfer learning accuracy on target datasets. Intriguingly, when we transfer networks trained on the combined ImageNet + iNat dataset and perform WS-DAN \cite{hu2019see} method over it, we got better results in Food-101 dataset. The resulted accuracy of the combination of ImageNet and iNat, fell in-between ImageNet and iNat pre-trained model. It means that we cannot attain good accuracy on target domains by just using a larger (combined) source domain. Our work demonstrates that a domain similarity score can be useful for identifying which large scale dataset to employ. That way, the model can learn essential features for the target dataset from large source training sets. Furthermore, we can employ attention aware data augmentation techniques to achieve state-of-the-art accuracy on several FGVC datasets.
\section{Conclusion}
In this paper, we describe a simple technique that takes attention mechanism as a data augmentation technique. Attention maps are guided to focus on the object’s parts and encourage multiple attention. We demonstrate that domain adaptive transfer learning plays a vital role in boosting performance. Depending on the domain similarity score, we can choose which source datasets to pre-train on to get better accuracy. We show that combining similarity-based selection of source datasets with attention-based augmentation technique can achieve state-of-the-art results in multiple fine-grained visual classification datasets. We also analyze the effect of image resolution
on transfer learning between the source and target domains. In future work, we are planning to explore the various factors on transfer learning to boost performance. We like to leverage variational auto encoder and GAN to generate augmented data which can be passed to the model to check the performance. Additionally, we want to compare different types of source datasets and try to control the variability in the number of training images to show the impact. 

\section*{Acknowledgement}
This work was partially supported by National Science Foundation grant IIS-1565328. Any opinions, findings, and conclusions or recommendations expressed in this publication are those of the authors, and do not necessarily reflect the views of the National Science Foundation.

\bibliographystyle{splncs04}
\bibliography{egbib}

\begin{thebibliography}{10}
\providecommand{\url}[1]{\texttt{#1}}
\providecommand{\urlprefix}{URL }
\providecommand{\doi}[1]{https://doi.org/#1}

\bibitem{abadi2016tensorflow}
Abadi, M., Barham, P., Chen, J., Chen, Z., Davis, A., Dean, J., Devin, M.,
  Ghemawat, S., Irving, G., Isard, M., et~al.: Tensorflow: A system for
  large-scale machine learning. In: 12th $\{$USENIX$\}$ Symposium on Operating
  Systems Design and Implementation ($\{$OSDI$\}$ 16). pp. 265--283 (2016)

\bibitem{anne2016deep}
Anne~Hendricks, L., Venugopalan, S., Rohrbach, M., Mooney, R., Saenko, K.,
  Darrell, T.: Deep compositional captioning: Describing novel object
  categories without paired training data. In: Proceedings of the IEEE
  conference on computer vision and pattern recognition. pp. 1--10 (2016)

\bibitem{azizpour2015factors}
Azizpour, H., Razavian, A.S., Sullivan, J., Maki, A., Carlsson, S.: Factors of
  transferability for a generic convnet representation. IEEE transactions on
  pattern analysis and machine intelligence  \textbf{38}(9),  1790--1802 (2015)

\bibitem{bao2017cvae}
Bao, J., Chen, D., Wen, F., Li, H., Hua, G.: Cvae-gan: fine-grained image
  generation through asymmetric training. In: Proceedings of the IEEE
  International Conference on Computer Vision. pp. 2745--2754 (2017)

\bibitem{cubuk2018autoaugment}
Cubuk, E.D., Zoph, B., Mane, D., Vasudevan, V., Le, Q.V.: Autoaugment: Learning
  augmentation policies from data. arXiv preprint arXiv:1805.09501  (2018)

\bibitem{cui2018large}
Cui, Y., Song, Y., Sun, C., Howard, A., Belongie, S.: Large scale fine-grained
  categorization and domain-specific transfer learning. In: Proceedings of the
  IEEE conference on computer vision and pattern recognition. pp. 4109--4118
  (2018)

\bibitem{deng2009imagenet}
Deng, J., Dong, W., Socher, R., Li, L.J., Li, K., Fei-Fei, L.: Imagenet: A
  large-scale hierarchical image database. In: 2009 IEEE conference on computer
  vision and pattern recognition. pp. 248--255. Ieee (2009)

\bibitem{donahue2014decaf}
Donahue, J., Jia, Y., Vinyals, O., Hoffman, J., Zhang, N., Tzeng, E., Darrell,
  T.: Decaf: A deep convolutional activation feature for generic visual
  recognition. In: International conference on machine learning. pp. 647--655
  (2014)

\bibitem{fu2017look}
Fu, J., Zheng, H., Mei, T.: Look closer to see better: Recurrent attention
  convolutional neural network for fine-grained image recognition. In:
  Proceedings of the IEEE conference on computer vision and pattern
  recognition. pp. 4438--4446 (2017)

\bibitem{ge2016fine}
Ge, Z., Bewley, A., McCool, C., Corke, P., Upcroft, B., Sanderson, C.:
  Fine-grained classification via mixture of deep convolutional neural
  networks. In: 2016 IEEE Winter Conference on Applications of Computer Vision
  (WACV). pp.~1--6. IEEE (2016)

\bibitem{he2016deep}
He, K., Zhang, X., Ren, S., Sun, J.: Deep residual learning for image
  recognition. In: Proceedings of the IEEE conference on computer vision and
  pattern recognition. pp. 770--778 (2016)

\bibitem{hu2019see}
Hu, T., Qi, H.: See better before looking closer: Weakly supervised data
  augmentation network for fine-grained visual classification. arXiv preprint
  arXiv:1901.09891  (2019)

\bibitem{khosla2011novel}
Khosla, A., Jayadevaprakash, N., Yao, B., Li, F.F.: Novel dataset for fgvc:
  Stanford dogs. In: San Diego: CVPR Workshop on FGVC. vol.~1 (2011)

\bibitem{krause20133d}
Krause, J., Stark, M., Deng, J., Fei-Fei, L.: 3d object representations for
  fine-grained categorization. In: Proceedings of the IEEE International
  Conference on Computer Vision Workshops. pp. 554--561 (2013)

\bibitem{krizhevsky2012imagenet}
Krizhevsky, A., Sutskever, I., Hinton, G.E.: Imagenet classification with deep
  convolutional neural networks. In: Advances in neural information processing
  systems. pp. 1097--1105 (2012)

\bibitem{li2018towards}
Li, P., Xie, J., Wang, Q., Gao, Z.: Towards faster training of global
  covariance pooling networks by iterative matrix square root normalization.
  In: Proceedings of the IEEE Conference on Computer Vision and Pattern
  Recognition. pp. 947--955 (2018)

\bibitem{lin2017improved}
Lin, T.Y., Maji, S.: Improved bilinear pooling with cnns. arXiv preprint
  arXiv:1707.06772  (2017)

\bibitem{lin2015bilinear}
Lin, T.Y., RoyChowdhury, A., Maji, S.: Bilinear cnn models for fine-grained
  visual recognition. In: Proceedings of the IEEE international conference on
  computer vision. pp. 1449--1457 (2015)

\bibitem{maji2013fine}
Maji, S., Rahtu, E., Kannala, J., Blaschko, M., Vedaldi, A.: Fine-grained
  visual classification of aircraft. arXiv preprint arXiv:1306.5151  (2013)

\bibitem{flowers}
Nilsback, M.E., Zisserman, A.: Automated flower classification over a large
  number of classes. In: 2008 Sixth Indian Conference on Computer Vision,
  Graphics \& Image Processing. pp. 722--729. IEEE (2008)

\bibitem{peng2018jointly}
Peng, X., Tang, Z., Yang, F., Feris, R.S., Metaxas, D.: Jointly optimize data
  augmentation and network training: Adversarial data augmentation in human
  pose estimation. In: Proceedings of the IEEE Conference on Computer Vision
  and Pattern Recognition. pp. 2226--2234 (2018)

\bibitem{rubner2000earth}
Rubner, Y., Tomasi, C., Guibas, L.J.: The earth mover's distance as a metric
  for image retrieval. International journal of computer vision
  \textbf{40}(2),  99--121 (2000)

\bibitem{sharif2014cnn}
Sharif~Razavian, A., Azizpour, H., Sullivan, J., Carlsson, S.: Cnn features
  off-the-shelf: an astounding baseline for recognition. In: Proceedings of the
  IEEE conference on computer vision and pattern recognition workshops. pp.
  806--813 (2014)

\bibitem{simon2015neural}
Simon, M., Rodner, E.: Neural activation constellations: Unsupervised part
  model discovery with convolutional networks. In: Proceedings of the IEEE
  International Conference on Computer Vision. pp. 1143--1151 (2015)

\bibitem{sun2017revisiting}
Sun, C., Shrivastava, A., Singh, S., Gupta, A.: Revisiting unreasonable
  effectiveness of data in deep learning era. In: Proceedings of the IEEE
  international conference on computer vision. pp. 843--852 (2017)

\bibitem{szegedy2016rethinking}
Szegedy, C., Vanhoucke, V., Ioffe, S., Shlens, J., Wojna, Z.: Rethinking the
  inception architecture for computer vision. In: Proceedings of the IEEE
  conference on computer vision and pattern recognition. pp. 2818--2826 (2016)

\bibitem{van2018inaturalist}
Van~Horn, G., Mac~Aodha, O., Song, Y., Cui, Y., Sun, C., Shepard, A., Adam, H.,
  Perona, P., Belongie, S.: The inaturalist species classification and
  detection dataset. In: Proceedings of the IEEE conference on computer vision
  and pattern recognition. pp. 8769--8778 (2018)

\bibitem{wah2011caltech}
Wah, C., Branson, S., Welinder, P., Perona, P., Belongie, S.: The caltech-ucsd
  birds-200-2011 dataset  (2011)

\bibitem{wang2018learning}
Wang, Y., Morariu, V.I., Davis, L.S.: Learning a discriminative filter bank
  within a cnn for fine-grained recognition. In: Proceedings of the IEEE
  Conference on Computer Vision and Pattern Recognition. pp. 4148--4157 (2018)

\bibitem{wen2016centerloss}
Wen, Y., Zhang, K., Li, Z., Qiao, Y.: A discriminative feature learning
  approach for deep face recognition. In: European conference on computer
  vision. pp. 499--515. Springer (2016)

\bibitem{xiao2015application}
Xiao, T., Xu, Y., Yang, K., Zhang, J., Peng, Y., Zhang, Z.: The application of
  two-level attention models in deep convolutional neural network for
  fine-grained image classification. In: Proceedings of the IEEE Conference on
  Computer Vision and Pattern Recognition. pp. 842--850 (2015)

\bibitem{yosinski2014transferable}
Yosinski, J., Clune, J., Bengio, Y., Lipson, H.: How transferable are features
  in deep neural networks? In: Advances in neural information processing
  systems. pp. 3320--3328 (2014)

\bibitem{yu2018deep}
Yu, F., Wang, D., Shelhamer, E., Darrell, T.: Deep layer aggregation. In:
  Proceedings of the IEEE Conference on Computer Vision and Pattern
  Recognition. pp. 2403--2412 (2018)

\bibitem{zhang2019survey}
Zhang, P., Zhong, Y., Deng, Y., Tang, X., Li, X.: A survey on deep learning of
  small sample in biomedical image analysis. arXiv preprint arXiv:1908.00473
  (2019)

\bibitem{zhang2018adversarial}
Zhang, X., Wei, Y., Feng, J., Yang, Y., Huang, T.S.: Adversarial complementary
  learning for weakly supervised object localization. In: Proceedings of the
  IEEE Conference on Computer Vision and Pattern Recognition. pp. 1325--1334
  (2018)

\bibitem{zheng2017learning}
Zheng, H., Fu, J., Mei, T., Luo, J.: Learning multi-attention convolutional
  neural network for fine-grained image recognition. In: Proceedings of the
  IEEE international conference on computer vision. pp. 5209--5217 (2017)

\end{thebibliography}

\end{document}